\documentclass[12pt]{article}

\usepackage{arabtex}

\usepackage{coling2018}

\usepackage{url}
\usepackage[para,online,flushleft]{threeparttable}
\usepackage{booktabs}
\usepackage{float}
\usepackage{hhline}
\usepackage{graphicx,subcaption}
\usepackage{smartdiagram}
\usepackage{pgf-pie}
\usepackage{caption} 

\usepackage{colortbl}

\usepackage[stable]{footmisc}

\usepackage{cp1256,utf8}
\setcode{utf8}

\date{}

\definecolor{lightgray}{rgb}{0.95,0.95,0.95}

\begin{document}

\title{Automatic Text Summarization (ATS) for Research Documents in Sorani Kurdish}

\author{
	\begin{tabular}[t]{c}
		Rondik Hadi Abdulrahman and Hossein Hassani\\
		\textnormal{University of Kurdistan Hewl\^er}\\
		\textnormal{Kurdistan Region - Iraq}\\
		{\tt {\{rondik.hadi, hosseinh}\}@ukh.edu.krd}
	\end{tabular}
}

\maketitle

\begin{abstract}
Extracting concise information from scientific documents aids learners, researchers, and practitioners. Automatic Text Summarization (ATS), a key Natural Language Processing (NLP) application, automates this process. While ATS methods exist for many languages, Kurdish remains underdeveloped due to limited resources. This study develops a dataset and language model based on 231 scientific papers in Sorani Kurdish, collected from four academic departments in two universities in the Kurdistan Region of Iraq (KRI), averaging 26 pages per document. Using Sentence Weighting and Term Frequency-Inverse Document Frequency (TF-IDF) algorithms, two experiments were conducted, differing in whether the conclusions were included. The average word count was 5,492.3 in the first experiment and 5,266.96 in the second. Results were evaluated manually and automatically using ROUGE-1, ROUGE-2, and ROUGE-L metrics, with the best accuracy reaching 19.58\%. Six experts conducted manual evaluations using three criteria, with results varying by document. This research provides valuable resources for Kurdish NLP researchers to advance ATS and related fields. 
\end{abstract}

\section{Introduction}
With the vast influx of information brought by the internet, Automatic Text Summarization (ATS) has emerged as a vital tool within Natural Language Processing (NLP). ATS distills complex, dense texts—such as scientific journal articles—into concise summaries that highlight key information \cite{Nikolov}. This alleviates the need for users to spend hours comprehending lengthy documents, making information more accessible \cite{Nazari}.

In today’s era, where quick access to information is essential, summarization enables the condensation of large volumes of text into meaningful summaries. Summarization techniques are categorized into two primary methods: extractive, which selects and combines sentences from the original text \cite{khademi}, and abstractive, which generates new sentences to convey the main ideas \cite{Daneshfar}. ATS can handle single-source texts or synthesize information from multiple sources, addressing diverse user needs \cite{Nenkova}. Furthermore, query-based models have been developed to produce summaries tailored to specific user queries \cite{Rahul}.

The overwhelming availability of information online underscores the need for efficient systems to extract critical insights quickly \cite{Rahul}. Automated summarization addresses this challenge by offering an alternative to manual methods, which are time-consuming and labor-intensive. Over the past decades, research has proposed and tested various ATS models on diverse datasets \cite{Rahul}.

Users of ATS tools range from casual readers seeking quick access to news to researchers navigating vast amounts of academic publications \cite{Elhadad}. Despite advancements, current ATS systems face challenges in natural fluency and deep understanding of unstructured text, requiring further development of intricate NLP techniques \cite{vilca}.

Text summarization involves multiple NLP tasks such as tokenization, stopword removal, and stemming \cite{Oliveira}. ATS research has evolved over five decades, translating linguistic theories into practical applications to enhance human-machine interaction \cite{Christian,Dutta}. The rise of big data has amplified the need for NLP tools, particularly in managing the abundance of textual content from sources like scientific literature, web platforms, and news articles \cite{Hunter}. ATS plays a pivotal role in filtering essential information from this content.

Machine learning approaches have been instrumental in ATS development, identifying features that help models generate accurate summaries. However, the effectiveness of these methods depends on the availability of suitable training corpora, which vary across 
languages and lack standardization \cite{Neto}.

The origins of Automatic Text Summarization (ATS) date back to the 1950s and 1960s, with \newcite{Luhn} pioneering work on "auto-abstracts," which relied on word frequency and distribution to score sentence importance. Later, \newcite{Edmundson1969NewMI} highlighted the limitations of frequency-based methods and introduced additional features such as cue words, title words, and sentence position to improve sentence selection.

In the 1970s and 1980s, AI-based summarization focused on rationalist techniques requiring hand-coded domain knowledge. These systems could produce summaries but were constrained by their dependence on domain-specific rules. By the 1990s, interest shifted away from such approaches, while Natural Language Generation (NLG) modules were paired with information extraction systems to produce more flexible summaries \cite{Orasan2019AutomaticS2}.

\begin{figure}[ht!]
	\centering
	\fbox{	\includegraphics[scale=0.6]{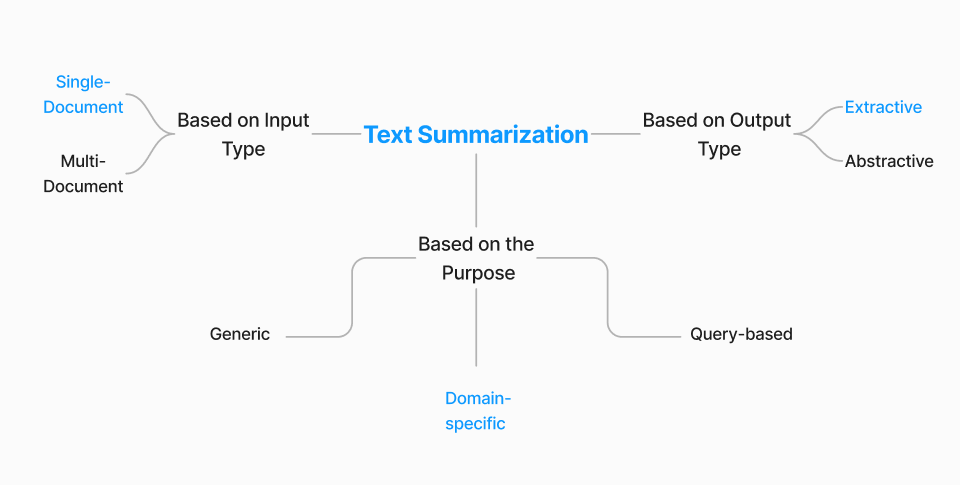}}
	\caption{Types of Text Summarization Approaches, adapted and modified from \cite{TextSummarization}.}
	\label{TextSummarization}
	
\end{figure}

Recent advancements in Natural Language Processing (NLP) and machine learning have led to the development of abstractive summarization, which mimics human summarization by understanding and rephrasing content. Although abstractive methods offer more natural summaries, they are computationally complex and require advanced NLP techniques \cite{Shafiee}. Figure \ref{TextSummarization} illustrates the ATS process, which is classified by input (single-document or multi-document), purpose (domain-specific, generic, or query-based), and output (extractive or abstractive). The choices made in this study are highlighted in blue.

This study focuses on ATS for Sorani Kurdish research documents. Kurdish, an Indo-European language spoken by over 30 million people, comprises several dialects, including Sorani, Kurmanji, and others \cite{Kurdishacademy,Hassani}. Sorani is one of the primary dialects spoken in Iraq and Iran, with Kurdish communities also present in regions like Turkey, Syria, and beyond \cite{Kreyenbroek}.

While text summarization research has been conducted for high- and low-resource languages such as English and Arabic, limited studies exist for Sorani Kurdish. This study aims to develop an ATS model tailored to Sorani Kurdish research documents. By employing NLP techniques, the proposed model will generate concise summaries, enabling Kurdish scholars to extract information efficiently. This work seeks to bridge language gaps in academic research, fostering collaboration and engagement within the Kurdish research community.

The Kurdish language remains low-resourced and suffers from necessary academic support despite the existence of different dialects and a substantially large speaker population, \cite{Hassani&kamal}. Furthermore, the language grapples with a lack of research initiatives, historical written texts, and comprehensive data collection activities \cite{Ahmadi}. Little research is conducted and readily available, and is concentrated within mostly in the Sorani dialect \cite{Hassani}. As the second most prevalent dialect, Sorani is primarily spoken among the 7 million individuals in the Kurdish communities across the regions of Iran and Iraq \cite{Ahmadi}.

The rest of this paper is organized into five sections: related  works, method, results and discussion, and conclusion.

\section{Related Work}
This section reviews previous research relevant to the study and is divided into two parts. The first part focuses on prior work related to Automatic Text Summarization (ATS) in the Kurdish language, while the second part covers research on summarization of scientific papers.

\subsection{ATS in Sorani Kurdish}

\newcite{BADAWI2023100043} introduces 'KurdSum,' recognized as the first benchmark dataset dedicated to text summarization in the Kurdish language, addressing a significant resource gap in Kurdish NLP. The KurdSum dataset is composed of news articles sourced from various Kurdish websites. Researchers utilized four extractive summarization models—LEXRANK, TEXTRANK, ORACLE, and LEAD-3—and three abstractive summarization models: Pointer-Generator, Sequence-to-Sequence, and Transformer-based approaches. The evaluation of abstractive methods revealed that the Pointer-Generator model achieved the highest ROUGE scores, surpassing other techniques. In extractive summarization, the ORACLE model delivered the best performance.

\newcite{Daneshfar} has developed an annotated corpus specifically for abstractive Kurdish text summarization. This corpus, compiled from various online news articles, includes concise summaries and serves as a foundation for developing and evaluating an abstractive text summarization model utilizing the mT5 transformer model. The dataset encompasses diverse categories such as health, politics, and sports. The research leverages embeddings for five languages, fine-tuned on the Kurdish corpus, and emphasizes the use of the mT5-base model. The model's performance is evaluated using ROUGE and N-gram novelty metrics, where it surpasses a baseline extractive method (Lead-3). 
 
\subsection{ATS in Scientific Papers}

\newcite{Slamet_2018} present a tool designed to automate the synthesis of Indonesian articles using TF-IDF and Vector Space Model (VSM) techniques. They organized their data before use by removing tables, figures, titles, and subtitles as well as abstracts of scientific articles. Their aim is to create abstracts that capture the essence of scientific articles and provide an effective alternative to manual synthesis. The system was tested using five Indonesian journal articles. The automatic abstracts generated by the system were compared with manually written abstracts to assess their similarity and effectiveness. The comparison was based on the presence of key phrases in both automatic and manual abstracts.

\newcite{Nikolov_Scientific} set forward a direction for text summarization using scientific articles both as a data basis for multi-sentence training and an inspiration source for document structure. They argue that the scientific paper format is structurally already containing summaries, in the form of the title and the abstract. They introduce two new datasets: title-gen, containing title-abstract pairs from biomed papers, and abstract-gen, containing abstract-body pairs from across domains. They evaluated a number of the extraction and abstraction neural compiled summarization approaches on these datasets. On the extraction side are the tfidf-emb and rwmd-rank approaches; on the abstraction side, there are the Long Short-Term Memory (LSTM), Convolutional Encoder-Decoder models (FConv), and Char-Level Encoder-Decoder models (C2C). The results indicate that the extractive approaches perform quite well, but the abstractive models, specifically fconv, perform best in the ability to generate coherent and relevant summaries.

\newcite{abu-jbara-radev-2011-coherent} introduce a method intended for generating readable, coherent summaries of scientific papers through citations. Their method utilizes posterior texts written by other researchers to underline the main aspects related to a target paper. Their work focuses on fluency, readability, and coherence of the summary. In this paper, the authors present an approach that applies three stages: preprocessing, extraction, and post-processing. They conduct experiments using the Association for Computational Linguistics (ACL) Anthology Network dataset for the evaluation of their approach. Results indicate huge gains over baseline systems, considering extraction quality and summary fluency. Their proposed approach obtained higher ROUGE scores and received better human ratings on coherence and readability compared to other techniques on summarization.

\newcite{Ronzano} explore how citation information can be used to generate more effective scientific article summaries. The authors conducted experiments using the BioSumm2014 dataset, which is a set of collections of scientific papers, where each collection contains a reference paper and papers that are citing the reference paper in their bibliographies. What they first and foremost want to find out is the effect of injecting citation contexts on the goodness of their summary. Their methodology generates relevant sentences from the scientific paper's sections like abstract, body, and citation context. Finally, all extracted sentences are ranked based on their ROUGE-2 score, which measures bigram overlap between generated and human reference summaries.

\section{Method}

We collect research documents from two universities in the Kurdistan Region of IRAQ. The texts are supposed to be in various forms, so we use different approaches and tools to convert them into plain texts. The data goes through several steps: cleansing, segmentation, tokenization, normalization, standardization, numeral unification, stemming, and stop word removal. Afterward, we apply the Sentence Weight and TF-IDF computations. Finally, the sentences are ranked in descending order based on their TF-IDF values to prioritize the most relevant content.

Once the data is ready, we create separate CSV files for training, validation, and testing of the models. In addition to the automated approach, we also perform a human-based evaluation in which we ask experts to rate the summarization output according to a five-point scale to evaluate the performance of our models.

\section{Results and Discussion}

\subsection{Data Collection}
We collected 231 research documents from four departments: Kurdish Language (66), Sociology (20), Social Sciences (44), and Political Sciences (101). Documents from the Kurdish Language, Sociology, and Social Sciences departments were obtained from the Zanco Academic Journal (ZJ) and the Lebanese French University (LFU) Journal. For Political Sciences, 55 documents were collected directly from the department (with permission), and 45 were sourced from the same journals. Access to ZJ was facilitated through the Salahaddin University Library, for which we were provided with a username and password. Table~\ref{tab:doc} summarizes the document counts, and average page numbers for each department.

\begin{table}[ht!]
	\centering
	\caption{Total Collected Data}
	\label{tab:doc}
	\begin{tabular}{|l|l|p{2.3cm}|p{1cm}|p{2cm}|}
		
		\hline
		\cellcolor{lightgray}{\textbf{\#}} & \cellcolor{lightgray}{\textbf{Department}} & \cellcolor{lightgray}{\textbf{Collected Documents}} & \cellcolor{lightgray}{\textbf{Avg pages}} & \cellcolor{lightgray}{\textbf{Domain Stop words}}  \\
		\hline
		1 & Political Sciences &  101 & 32.5 & 22 \\
		\hline
		2 & Kurdish Language &  66 & 23.1 & 18 \\
		\hline
		3 & Sociology &  44 & 28.4 & 30 \\
		\hline
		4 & Social Sciences &  20 & 30.4 & 6 \\
		\hline
		* & Total &  231 & 26 & 76 \\
		\hline
		
	\end{tabular}
	
\end{table}

Additionally, domain-specific stop words were compiled for each department through expert consultation and online research, then validated by experts and saved in JSON files within the 'domain\_stopwords' folder, later used in the model. 

\subsection{Data Conversion}
Converting Kurdish language PDFs to text was challenging due to the lack of reliable OCR tools. We tried various OCR systems, including i2OCR, PDF2Doc, Adobe Reader, Foxit, and ABBYY, with mixed results depending on the document type. Saman Idrees assisted  in converting 11 documents based on their work on Kurdish OCR \cite{Saman}. Foxit PDF Reader provided the best overall results, though some PDFs had encoding and font issues. To fix errors like extra line breaks, incorrect character rendering, and text alignment (right-to-left), we developed Python scripts. The entire process took around two months, involving manual text verification, though the final documents are not 100\% error-free.

\subsection{Data Cleaning}
After converting the documents, we cleaned the text by removing abstracts, references, tables, figures, side texts, headings, and titles, focusing solely on the body text for summarization. We noticed inconsistent formatting, punctuation errors, mixing of Arabic and English numerals, and variations in spelling of the same word (e.g., ``{\small{\<کەلپور>}} '' and ``{\small{\<کەلپوور>}}~'').
However, we kept the original text to preserve document integrity. Each document took about 20 minutes to clean, followed by a final check for accuracy.

\subsection{Text Preprocessing}
To ensure preprocessing accuracy, we executed the script after each step and reviewed the outputs. Results from each stage were saved in the 'process' folder. Table~\ref{tab:pre-output} shows the outcomes for the sample sentence: 
``{\small \RL{ڕوسیا وەک ولاتێکی کاریگەریی خاوەن هێز کە توانایەکی زۆری هەیە و ئەتوانێت لە دۆز و هاوکێشە نێودەوڵەتیەکان ڕۆڵی گەورە بگێرێت،}}~''. 
The 'Processed\_doc.txt' file captures the text after tokenization, normalization, standardization, numeral unification, stemming, and stop word removal. The 'Processed\_doc.xml' file stores the XML version for further analysis. To verify stop word removal, 'Debug\_doc.txt' lists detected stop words and their count, confirming the removal of domain-specific stop words. The 'Processed\_doc\_tokens.txt' file displays all tokens, each on a separate line, to validate tokenization. We ensured script reliability by testing after each modification before moving to the next stage. We used the Punkt algorithm, which performs appropriately on Kurdish \cite{Roshna}, on our corpus for accurate segmentation. Next, we extracted features, converting the preprocessed text into numerical embeddings capturing semantic and contextual meaning. The segmentation and feature extraction results were saved as PKL files in the 'segmenters' and 'features' folders. In each run, the model performs these steps to refine its performance.

\begin{table}[ht!]
    \centering
    \caption{Preprocessing output sample for the document 'doc'}
\label{tab:pre-output}
  \begin{tabular}{|l|l|p{6cm}|}

     \hline
     \cellcolor{lightgray}{\textbf{\#}} & \cellcolor{lightgray}{\textbf{Output file}} & \cellcolor{lightgray}{\textbf{Content}}  \\
    \hline
     1 & Processed\_doc1.txt & {\small \RL{ڕوسیا \_ ولات \_کاریگەر\_ \_خاوەن \_هێز \_ \_توانایەک \_زۆر \_هەیە\_ توان \_دۆز\_ \_هاوکێشە\_ \_نێودەوڵەتی\_ ڕۆڵ \_گەورە\_ بگێرێ ، }} 
     \\
    \hline
    2 & Processed\_doc1.xml & <s> {\small \RL{ڕوسیا \_ ولات \_کاریگەر\_ \_خاوەن \_هێز \_ \_توانایەک \_زۆر \_هەیە\_ توان \_دۆز\_ \_هاوکێشە\_ \_نێودەوڵەتی\_ ڕۆڵ \_گەورە\_ بگێرێ ،...}} \verb|<\s>| 
    \\
    \hline
    3 & Debug\_doc1.txt &  Stop words removed one by one: ({\small \RL{وەک}} , {\small \RL{کە}} , {\small \RL{لە}} ,...), 
    Number of stop words removed: 2375 \\
    \hline
    4 & Processed\_doc1\_tokens.txt &  Line 1: \_{\small \RL{\<ڕوسیا>}}\_ , Line 2: \_{\small \RL{ولات}}\_ , Line 3: \_{\small \RL{کاریگەر}}\_ {,...}
    \\
    \hline
     
\end{tabular}

\end{table}

\subsection{Developing the Model} 

After applying the Sentence Weight Algorithm, the script generated an output file named ``Processed\_Sentence\_Weight\_doc.txt,'' verifying that the algorithm’s calculations were executed correctly. Table~\ref{tab:Apre-output} displays the sentence weighting results for the sample sentence: {``{\small {\<هەوڵ بدرێت سەرجەم پێکدادان و تێهەڵچوونەکان لە چوارچێوەی وڵاتانی یەکیەتی سۆڤیەتدا پێشوو>}}~''}. 
Each sentence in the document was assigned a weight between 0 and 1, indicating its relative importance within the text.

After obtaining these weights, we eliminated the bottom 50\% of sentences with the lowest scores to retain the most significant content. We then applied the TF-IDF algorithm to evaluate word importance in each remaining sentence. Table~\ref{tab:Apre-output} shows the TF-IDF score for the sample sentence, with the complete results saved in `'Processed\_TF-IDF\_doc.txt'' for verification.

Finally, the sentences were ranked in descending order based on their TF-IDF values to prioritize the most relevant content. The sorted output was saved in 'Sorted\_TF-IDF\_doc.txt,' confirming that the sorting process was performed accurately.

 \begin{table}[ht!]
    \centering
    \caption{The post-processing output sample for the document 'doc'}
\label{tab:Apre-output}
  \begin{tabular}{|l|l|p{6cm}|}

     \hline
     \cellcolor{lightgray}{\textbf{\#}} & \cellcolor{lightgray}{\textbf{Output file}} & \cellcolor{lightgray}{\textbf{Content}}  \\
    \hline
     1 & Processed\_Sentence\_Weight\_doc.txt & Sentence: {\small \RL{هەوڵ \_ بدرێت \_سەرجەم\_ \_پێک \_دادان \_ \_تێ-هەڵ-چوون \_ەکان \_لە\_ چێو \_وڵات\_ \_ان\_ \_یەکیەت\_ سۆڤیەت \_پێ\_ شوو . }} 
     Weight: 0.441 \\
    \hline
     2 & Processed\_TF-IDF\_doc.txt & Sentence: {\small \RL{هەوڵ \_ بدرێت \_سەرجەم\_ \_پێک \_دادان \_ \_تێ-هەڵ-چوون \_ەکان \_لە\_ چێو \_وڵات\_ \_ان\_ \_یەکیەت\_ سۆڤیەت \_پێ\_ شوو . }} 
     TF-IDF Weight: 0.002 \\
    \hline
     
\end{tabular}

 \end{table}

 As we decided to evaluate our summary results against the abstracts of the documents, we first examined the length of the abstracts across all documents. Using a Python script, we calculated the average word count for each department's abstracts and observed considerable variation. As shown in table~\ref{tab:avg}, the average word count per department is presented alongside the overall average, which is approximately 182 words. Based on this analysis, we decided to limit the length of our summaries.
 
 \begin{table}[ht!]
    \centering
    \caption{Average Abstract Word Count by Department and Overall}
\label{tab:avg}
  \begin{tabular}{|l|p{3.5cm}|p{3cm}|}

     \hline
     \cellcolor{lightgray}{\textbf{\#}} & \cellcolor{lightgray}{\textbf{Department}} & \cellcolor{lightgray}{\textbf{Abstract Avg. Word Count}}  \\
    \hline
     1 & Political Sciences &  207.42 \\
    \hline
    2 & Kurdish Language &  154.26 \\
    \hline
    3 & Sociology & 180.09  \\
    \hline
    4 & Social Sciences & 184.1  \\
    \hline
    * & Average &  181.46 \\
    \hline
     
\end{tabular}

 \end{table}
 
The word count of our summaries varied depending on the structure and length of the input documents. Documents with longer or more complex structures naturally produced more sentences, potentially resulting in longer summaries. Additionally, if the sentences in some documents were significantly longer or shorter than in others, this directly impacted the word count of the resulting summaries.

\subsection{Evaluation}

After completing the development and training of our model for ATS, we proceeded to evaluate its performance across various departments. The results were systematically saved for each department, ensuring that the outputs from different stages of the process—such as training, validation, and testing—were clearly delineated. This separation of summaries by stage enabled us to assess the model's consistency and reliability. The model produced two primary types of summaries: a full summarization of the document and a more concise, final summary with a limited word count. Additionally, for each document, we generated a summary state file output. This output detailed the number of sentences included in the summary, the word count for each sentence, the total word count, and the total number of sentences. This level of detail provided further insights into the structure and effectiveness of the generated summaries.

W developed a Python script to prepare the dataset the mentioned domains (Kurdish Language, Political Sciences, Social Sciences, and Sociology) demonstrating its broad applicability. The script processes documents and summaries for each domain, creating separate CSV files for training, validation, and testing.

Due to the relatively small dataset (231 documents), we allocated 15\% for testing, with the remaining 85\% split into approximately 71\% for training and 14\% for validation. This results in 71 training, 15 validation, and 15 testing documents for Political Science; 46, 10, and 10 for Kurdish Language; 14, 3, and 3 for Social Science; and 30, 7, and 7 for Sociology. This ensures sufficient training data while reserving unseen data for evaluation, improving the model’s generalizability.

\subsection{Experiments}

We conducted two experiments on the available data. In the first experiment, the documents included their research conclusions, while in the second, these conclusions were excluded. This approach was taken to assess how the presence of conclusions would influence the summarization results. We have previously explained and clarified the number of documents for each department. We evaluated our results in both experiments. The difference between the average word count for each department individually and the combined average across all departments, which is 225.34 words, is shown in table~\ref{tab:avg_word}.

\begin{table}[ht!]
    \centering
    \caption{Difference in Average Word Count Between Documents With and Without a Conclusion}
\label{tab:avg_word}
  \begin{tabular}{|l|l|p{2cm}|p{2.5cm}|}
     \hline
     \cellcolor{lightgray}{\textbf{\#}} & \cellcolor{lightgray}{\textbf{Department}} & \cellcolor{lightgray}{\textbf{Avg word with Con.}} & \cellcolor{lightgray}{\textbf{Avg word without Con.}}   \\
    \hline
     1 & Political Sciences &  6179.22 & 5892.82 \\
    \hline
    2 & Kurdish Language & 5840.98 & 5654.64  \\
    \hline
    3 & Sociology &  5383.37 & 5146.65  \\
    \hline
    4 & Social Sciences &  4565.65 & 4373.75 \\
    \hline
    * & Average &  5492.3 & 5266.96 \\
    \hline
\end{tabular}

 \end{table}

\subsubsection{First Experiment}

The first experiment was conducted on documents that included their conclusions. The data were randomly divided into training, validation, and testing sets. For each document, we obtained summary results, which were then organized into three separate folders corresponding to training, validation, and testing. This separation allowed us to easily identify which documents were selected for each stage—training, validation, or testing. In a separate folder, the model stored the evaluation results for each document. Additionally, the model calculated and stored the average evaluation for all the documents within each department. The results for Precision, Recall, and F-Measure across all three ROUGE metrics—ROUGE-1, ROUGE-2, and ROUGE-L—are presented in table~\ref{tab:rouge_conclusion}.

\begin{table}[ht!]
    \centering
        \caption{ROUGE Evaluation Result with Conclusion}
    \label{tab:rouge_conclusion}
    \begin{tabular}{|l|l|l|l|}
        \hline
        \cellcolor{lightgray}{\textbf{Departments}} &
        \multicolumn{3}{|c|}{\cellcolor{lightgray}{\textbf{ROUGE-1}}} \\
        \hhline{*{2}{~}{{----}}}
        \cline{2-4}
        \cellcolor{lightgray} & 
        \multicolumn{1}{c|}{\cellcolor{lightgray}{\textbf{Precision}}} & \multicolumn{1}{c|}{\cellcolor{lightgray}{\textbf{Recall}}} & \multicolumn{1}{c|}{\cellcolor{lightgray}{\textbf{F-Score}}}\\
        \hline
        \cellcolor{lightgray} Political Science  & 0.1267 & 0.1741 &  0.1410  \\
        \hline
        \cellcolor{lightgray} Kurdish Language & 0.1162 &  0.1726 &  0.1347  \\
        \hline
        \cellcolor{lightgray} Sociology & 0.1296 & 0.1645 &  0.1436  \\
        \hline
        \cellcolor{lightgray} Social Science & 0.1176 & 0.1948 & 0.1462   \\
        \hline
        \cellcolor{lightgray} Average & 0.122525 & 0.1765 & 0.141375 \\
        \hline
        
        \cellcolor{lightgray}{\textbf{Departments}} &
        \multicolumn{3}{|c|}{\cellcolor{lightgray}{\textbf{ROUGE-2}}} \\
        \hhline{*{2}{~}{{----}}}
        \cline{2-4}
        \cellcolor{lightgray} & 
        \multicolumn{1}{c|}{\cellcolor{lightgray}{\textbf{Precision}}} & \multicolumn{1}{c|}{\cellcolor{lightgray}{\textbf{Recall}}} & \multicolumn{1}{c|}{\cellcolor{lightgray}{\textbf{F-Score}}}\\
        \hline
        \cellcolor{lightgray} Political Science  & 0.0260 & 0.0342 &  0.0285  \\
        \hline
        \cellcolor{lightgray} Kurdish Language & 0.0186 & 0.0239 &  0.0203  \\
        \hline
        \cellcolor{lightgray} Sociology & 0.0317 & 0.0339 &  0.0323  \\
        \hline
        \cellcolor{lightgray} Social Science & 0.0205 & 0.0349 &  0.0257  \\
        \hline
        \cellcolor{lightgray} Average & 0.0242 & 0.031725 & 0.0267 \\
        \hline
        
        \cellcolor{lightgray}{\textbf{Departments}} &
        \multicolumn{3}{|c|}{\cellcolor{lightgray}{\textbf{ROUGE-L}}} \\
        \hhline{*{2}{~}{{----}}}
        \cline{2-4}
        \cellcolor{lightgray} & 
        \multicolumn{1}{c|}{\cellcolor{lightgray}{\textbf{Precision}}} & \multicolumn{1}{c|}{\cellcolor{lightgray}{\textbf{Recall}}} & \multicolumn{1}{c|}{\cellcolor{lightgray}{\textbf{F-Score}}}\\
        \hline
        \cellcolor{lightgray} Political Science  & 0.0881 & 0.1242 &  0.0993  \\
        \hline
        \cellcolor{lightgray} Kurdish Language & 0.0735 & 0.1105 &  0.0853  \\
        \hline
        \cellcolor{lightgray} Sociology & 0.1082 & 0.1343 &  0.1188  \\
        \hline
        \cellcolor{lightgray} Social Science & 0.0980 & 0.1613 &  0.1215  \\
        \hline
        \cellcolor{lightgray} Average & 0.09195 & 0.132575 & 0.106225 \\
        \hline
    \end{tabular}

\end{table}

\begin{figure}[ht!]
 		\centering
		\fbox{	\includegraphics[scale=0.7]{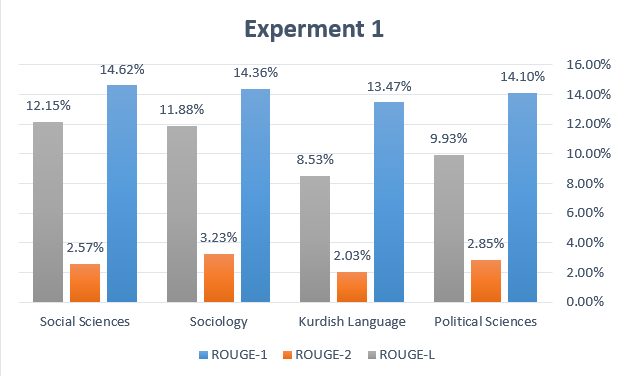}}
		\caption{Visual representation of Experiment 1 results.}
		\label{Ep1}
		
	\end{figure}

As observed in the results, ROUGE-1 yielded the best performance among the three metrics for each department. Additionally, the average ROUGE-1 score across all departments was the highest, while the ROUGE-2 score was the lowest among the three ROUGE metrics. Figure~\ref{Ep1} provides a visual representation of the differences in scores, highlighting the model’s effectiveness in capturing lexical overlap (ROUGE-1), bi-gram overlap (ROUGE-2), and longest common subsequences (ROUGE-L) across the four departments.

\subsubsection{Second Experiment}

In the second experiment, we followed the same process as in the first experiment and obtained the results in the same manner. The key difference lies in the data, as the documents in this experiment did not include the research conclusions. The results for the three ROUGE metrics for this experiment can be seen in table~\ref{tab:rouge_withou-conclusion}.

\begin{table}[ht!]
    \centering
        \caption{ROUGE Evaluation Result without Conclusion}
    \label{tab:rouge_withou-conclusion}
    \begin{tabular}{|l|l|l|l|}
        \hline
        \cellcolor{lightgray}{\textbf{Departments}} &
        \multicolumn{3}{|c|}{\cellcolor{lightgray}{\textbf{ROUGE-1}}} \\
        \hhline{*{2}{~}{{----}}}
        \cline{2-4}
        \cellcolor{lightgray} & 
        \multicolumn{1}{c|}{\cellcolor{lightgray}{\textbf{Precision}}} & \multicolumn{1}{c|}{\cellcolor{lightgray}{\textbf{Recall}}} & \multicolumn{1}{c|}{\cellcolor{lightgray}{\textbf{F-Score}}}\\
        \hline
        \cellcolor{lightgray} Political Science  & 0.1240 & 0.1716 &  0.1375  \\
        \hline
        \cellcolor{lightgray} Kurdish Language & 0.0937 & 0.1692 &  0.1169  \\
        \hline
        \cellcolor{lightgray} Sociology & 0.1459 & 0.2228 &  0.1735 \\
        \hline
        \cellcolor{lightgray} Social Science & 0.1578 & 0.2659 & 0.1958 \\
        \hline
        \cellcolor{lightgray} Average & 0.13035 & 0.207375 & 0.155925 \\
        \hline

        \cellcolor{lightgray}{\textbf{Departments}} &
        \multicolumn{3}{|c|}{\cellcolor{lightgray}{\textbf{ROUGE-2}}} \\
        \hhline{*{2}{~}{{----}}}
        \cline{2-4}
        \cellcolor{lightgray} & 
        \multicolumn{1}{c|}{\cellcolor{lightgray}{\textbf{Precision}}} & \multicolumn{1}{c|}{\cellcolor{lightgray}{\textbf{Recall}}} & \multicolumn{1}{c|}{\cellcolor{lightgray}{\textbf{F-Score}}}\\
        \hline
        \cellcolor{lightgray} Political Science  & 0.0208 & 0.0268 &  0.0224  \\
        \hline
        \cellcolor{lightgray} Kurdish Language & 0.0100 & 0.0156 &  0.0121  \\
        \hline
        \cellcolor{lightgray} Sociology & 0.0482 & 0.0672 &  0.0545 \\
        \hline
        \cellcolor{lightgray} Social Science & 0.0487 & 0.0818 & 0.0603 \\
        \hline
        \cellcolor{lightgray} Average & 0.031925 & 0.04785 & 0.037325 \\
        \hline
        
        \cellcolor{lightgray}{\textbf{Departments}} &
        \multicolumn{3}{|c|}{\cellcolor{lightgray}{\textbf{ROUGE-L}}} \\
        \hhline{*{2}{~}{{----}}}
        \cline{2-4}
        \cellcolor{lightgray} & 
        \multicolumn{1}{c|}{\cellcolor{lightgray}{\textbf{Precision}}} & \multicolumn{1}{c|}{\cellcolor{lightgray}{\textbf{Recall}}} & \multicolumn{1}{c|}{\cellcolor{lightgray}{\textbf{F-Score}}}\\
        \hline
        \cellcolor{lightgray} Political Science  & 0.0858 & 0.1214 &  0.0961  \\
        \hline
        \cellcolor{lightgray} Kurdish Language & 0.0690 & 0.1268 & 0.0867 \\
        \hline
        \cellcolor{lightgray} Sociology & 0.1239 & 0.1876 &  0.1466 \\
        \hline
        \cellcolor{lightgray} Social Science & 0.1304 & 0.2189 &  0.1616  \\
        \hline
        \cellcolor{lightgray} Average & 0.102275 & 0.163675 & 0.12275 \\
        \hline
    \end{tabular}

\end{table}

\begin{figure}[ht!]
 		\centering
		\fbox{	\includegraphics[scale=0.8]{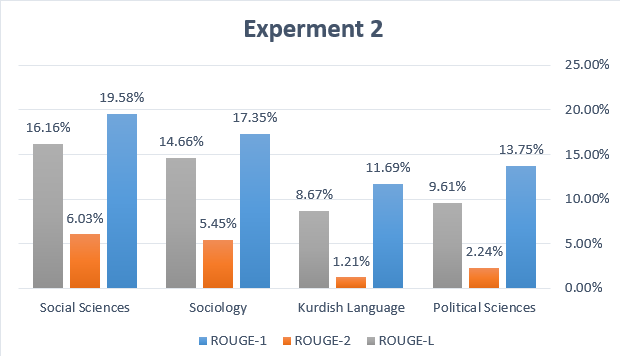}}
		\caption{Visual representation of Experiment 2 results.}
		\label{Ep2}
		
	\end{figure}

\begin{figure}[ht!]
 		\centering
		\fbox{	\includegraphics[scale=0.42]{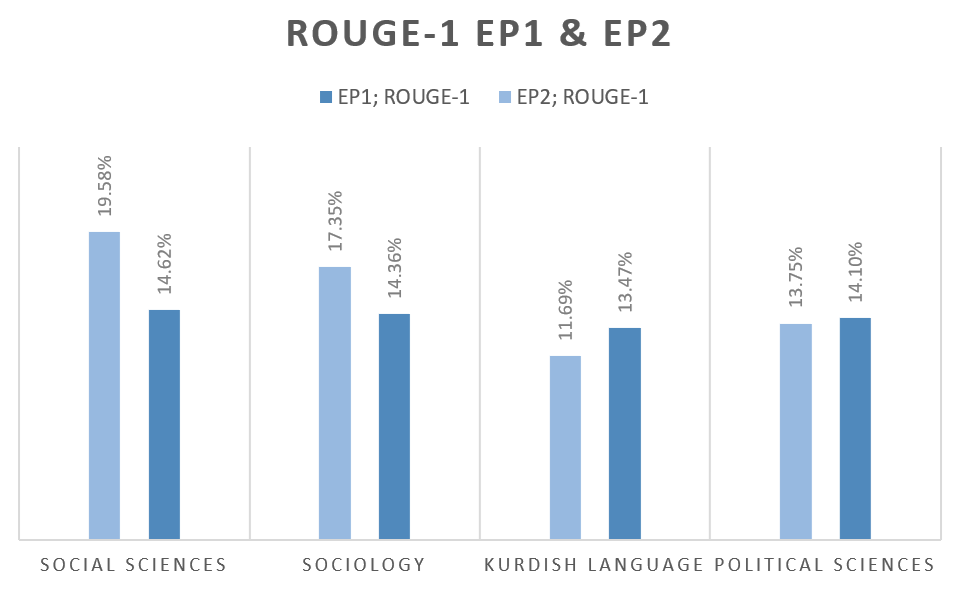}}
		\caption{Comparison of ROUGE-1 results from both experiments.}
		\label{Ep1&Ep2}
		
	\end{figure}

As shown in the results, similar to the first experiment, ROUGE-1 yielded the best performance among the three metrics for each department. Once again, the average ROUGE-2 score across all departments was the lowest among the three ROUGE metrics. Figure~\ref{Ep2} illustrates the score variations in Experiment 2, showcasing the model's performance in ROUGE-1, ROUGE-2, and ROUGE-L across the four departments, while figure~\ref{Ep1&Ep2} compares the ROUGE-1 results of both experiments.

Two separate experiments were conducted to evaluate our model's performance, each involving the processes of training, validation, and testing on its respective dataset. Both experiments were performed on the same server, equipped with 146 GB of RAM and 46 cores. In the first experiment, which included research papers with their conclusion sections, the training process took approximately 22 hours. In the second experiment, which excluded these conclusions, training took around 20 hours under similar conditions. For both experiments, the data was split into 70\% for training, 15\% for validation, and 15\% for testing. 

  \begin{table}[ht!]
    \centering
    \caption{Comparison of Experimental Setups and Results}
\label{comp}
  \begin{tabular}{|p{2.1cm}|p{3.1cm}|p{3.1cm}|p{3.1cm}|}

     \hline
      \cellcolor{lightgray}{\textbf{Feature}} & \cellcolor{lightgray}{\textbf{Experiment 1}} & \cellcolor{lightgray}{\textbf{Experiment 2}} & \cellcolor{lightgray}{\textbf{Remarks
}}  \\
    \hline
     \cellcolor{lightgray} Dataset & Research papers including conclusion sections & Research papers excluding conclusion sections & Difference in inclusion of conclusion sections. \\
    \hline
     \cellcolor{lightgray} Training Time	 &  Approximately 22 hours & Approximately 20 hours & Slightly longer training time for Experiment 1.
 \\
    \hline
     \cellcolor{lightgray} Server Configuration &  146 GB RAM, 46 cores & 146 GB RAM, 46 cores & Same server setup for both experiments.  \\
    \hline
    \cellcolor{lightgray} Data Split Ratio &  70\% training, 15\% validation, 15\% testing & 70\% training, 15\% validation, 15\% testing & Identical data splits used in both experiments. \\
    \hline
     \cellcolor{lightgray} Best Evaluation Result & 14.62\%(ROUGE-1, Social Science department) & 19.58\%(ROUGE-1, Social Science department) & Both results achieved in the same evaluation metric and domain.  \\
    \hline
     
\end{tabular}

 \end{table}
 
\subsection{Human-based Evaluation}
This section presents the manual evaluation conducted by six experts, who assessed the summaries by completing three evaluation forms that we prepared specifically for this purpose.

\subsubsection{First Approach}
\begin{figure}[ht!]
 		\centering
		\fbox{	\includegraphics[scale=0.6]{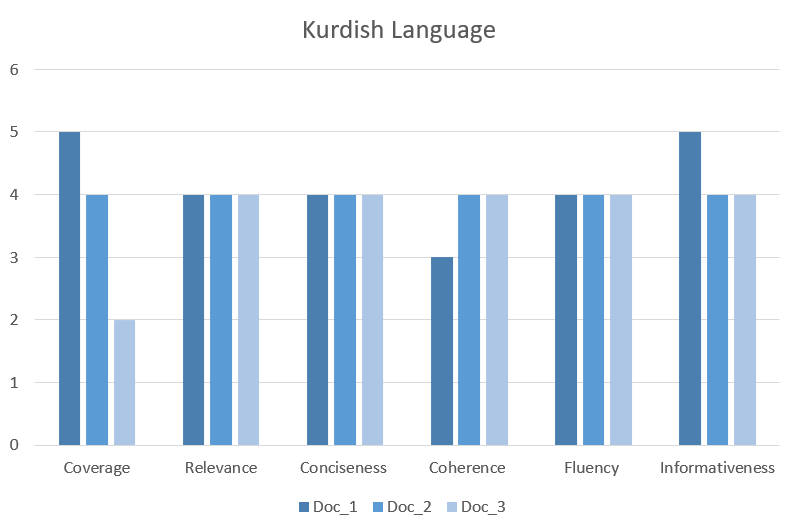}}
		\caption{Comparison of Evaluator Feedback for the Kurdish Language Department Document }
		\label{kurdish}
		
	\end{figure}

\begin{figure}[ht!]
 		\centering
		\fbox{	\includegraphics[scale=0.6]{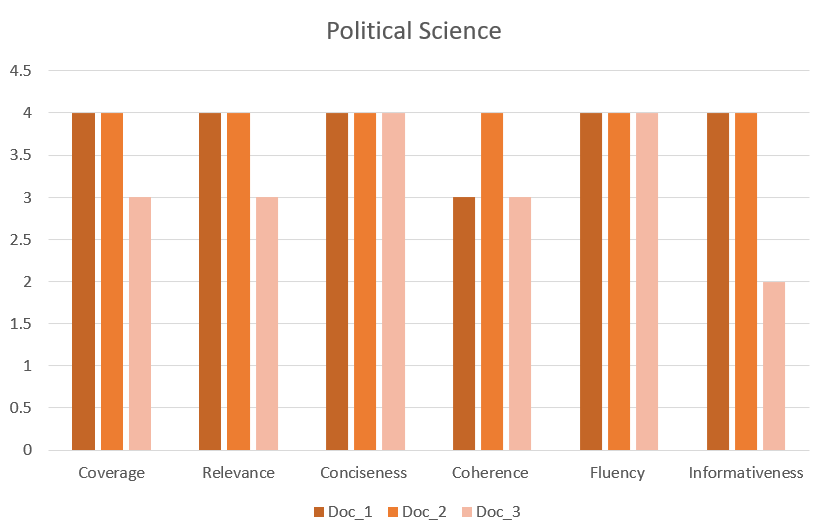}}
		\caption{Comparison of Evaluator Feedback for the Political Science Department Document  }
		\label{politic}
		
	\end{figure}

This manual evaluation involved asking domain experts to compare the abstract of the research documents with our generated summaries. The evaluation was conducted on documents from the 'Kurdish Language' and 'Political Science' departments. A selection of documents from each department was prepared for expert review. We collaborated with two evaluators who provided expertise in Kurdish language and Political Law. Both evaluators reviewed and rated the documents, contributing to the assessment of our results. Figures \ref{kurdish} and \ref{politic} visualize these results in chart form. The results are based on the experts' responses to a structured evaluation form that we provided. Each question on the form was rated on a five-point scale: very poor, poor, fair, good, and excellent.

\subsubsection{Second Approach}

\begin{figure}[H]
 		\centering
		\fbox{	\includegraphics[scale=0.6]{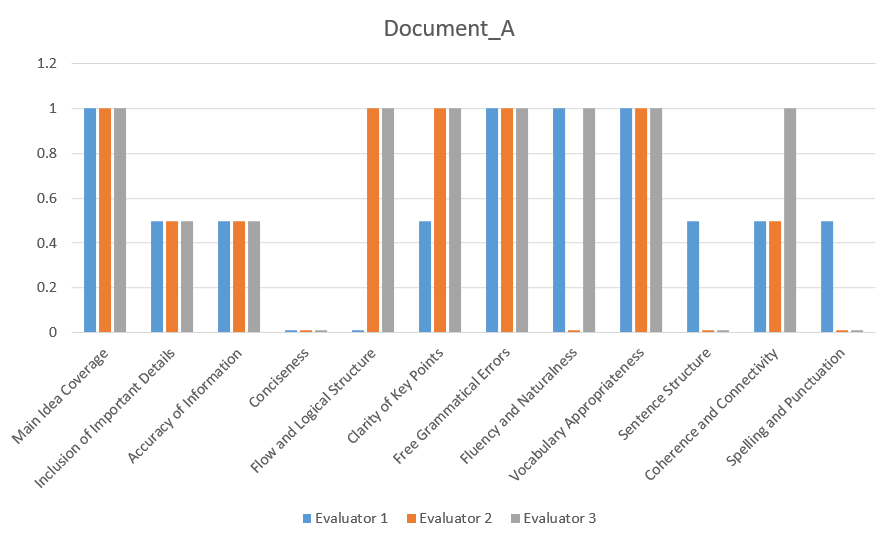}}
		\caption{Comparison of Evaluator Feedback for Doc\_A Second Evaluation.}
		\label{A_2}
		
	\end{figure}

\begin{figure}[H]
 		\centering
		\fbox{	\includegraphics[scale=0.6]{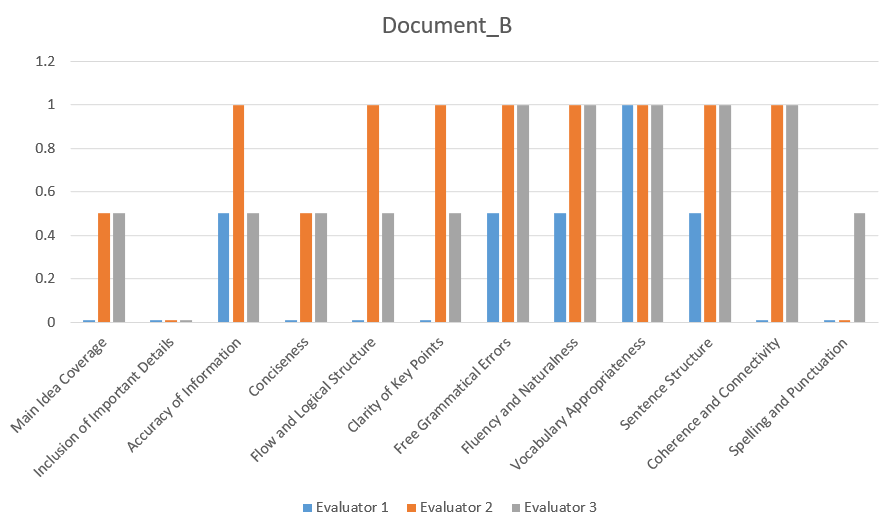}}
		\caption{Comparison of Evaluator Feedback for Doc\_A Second Evaluation.}
		\label{B_2}
		
	\end{figure}

\begin{figure}[H]
 		\centering
		\fbox{	\includegraphics[scale=0.6]{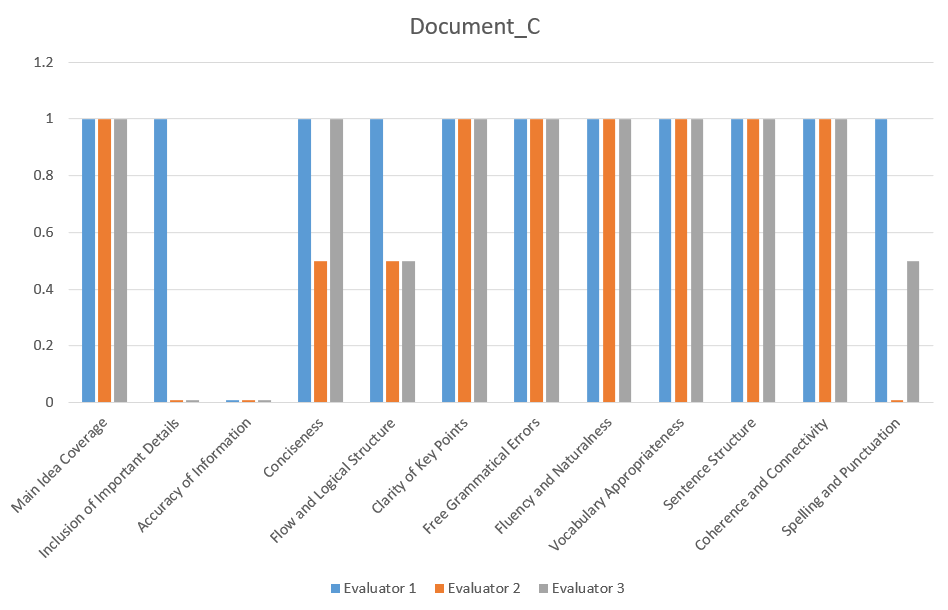}}
		\caption{Comparison of Evaluator Feedback for Doc\_A Second Evaluation.}
		\label{C_2}
		
	\end{figure}

A second evaluation form, available in both Kurdish and English, was designed to address concerns raised in the initial evaluation. Unlike the first form, which focused on rating scales, this version emphasized detailed feedback by comparing generated summaries with research abstracts. It included 12 questions—six assessing content quality and six evaluating grammatical accuracy. The content questions examined whether the summary captured the main idea, included key details, maintained factual accuracy, was concise, logically structured, and easy to understand. The grammar questions checked for errors, fluency, appropriate vocabulary, proper sentence structure, coherence, and spelling or punctuation issues. Figures \ref{A_2}, \ref{B_2}, and \ref{C_2} visualize the results of this evaluation in chart form. These results are based on the experts’ responses to the structured evaluation form. The chart represents three evaluation stages: 0, 0.5, and 1. A score of 0 indicates a negative response, suggesting the evaluator identified shortcomings in the summary based on the 12 criteria. A score of 0.5 reflects a neutral stance, indicating a mix of positive and negative aspects. A score of 1 represents a positive response. The evaluation was conducted by three experts, and the scores are reported for three documents, referred to as Document A, Document B, and Document C.

\subsubsection{Third Approach}
\begin{figure}[H]
 		\centering
		\fbox{	\includegraphics[scale=0.6]{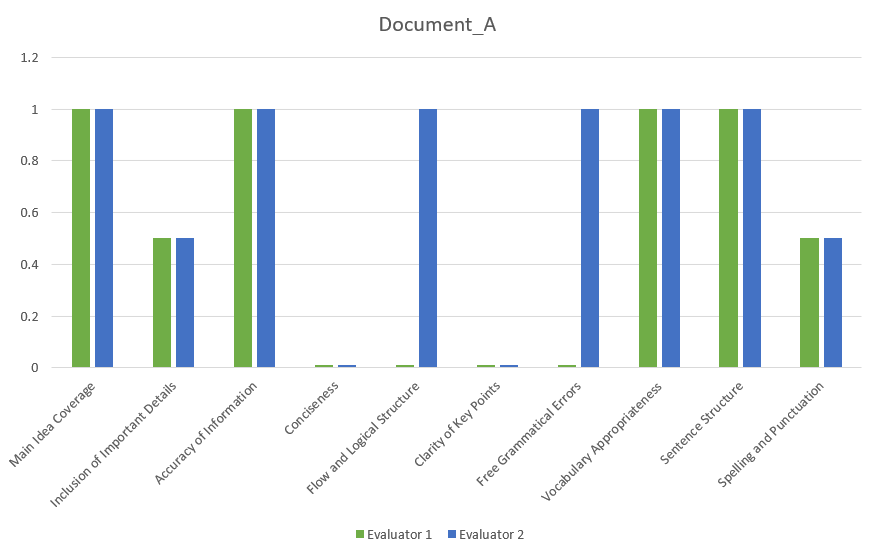}}
		\caption{Comparison of Evaluator Feedback for Doc\_B Third Evaluation.}
		\label{A_3}
		
	\end{figure}

\begin{figure}[H]
 		\centering
		\fbox{	\includegraphics[scale=0.6]{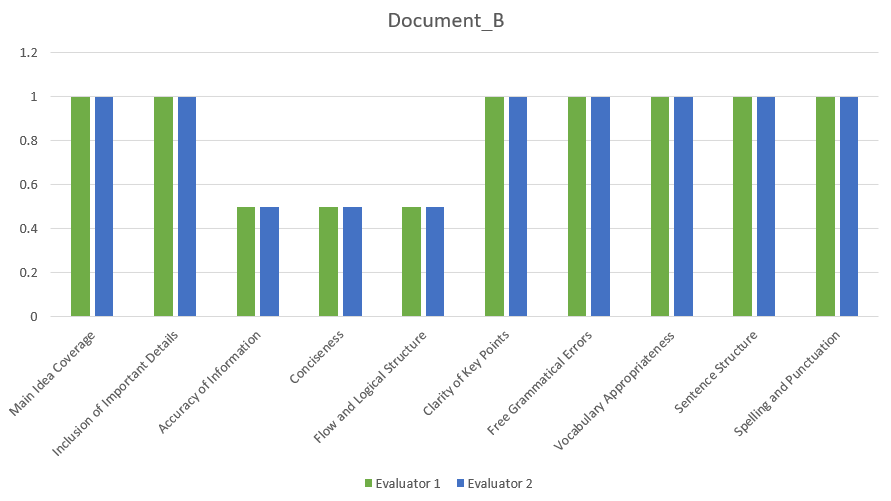}}
		\caption{Comparison of Evaluator Feedback for Doc\_B Third Evaluation.}
		\label{B_3}
		
	\end{figure}

In this evaluation, to ensure the accuracy of the generated summaries, evaluators first read the full research document, then reviewed the generated summary, and finally completed an evaluation form. This approach ensured that the evaluation was based on the document itself, not influenced by its abstract. The form, prepared in both Kurdish and English, contained ten open-ended questions covering content quality and grammatical accuracy. It assessed whether the summary captured the main ideas, included essential details, maintained factual accuracy, was concise, and had a logical flow. It also checked for grammatical errors, appropriate vocabulary, proper sentence structure, and any spelling or punctuation issues. Figures \ref{A_3} and \ref{B_3} visualize the results of this evaluation in chart form. These results are based on the experts’ responses to the structured evaluation form. The chart represents three evaluation stages: 0, 0.5, and 1. A score of 0 indicates a negative response, meaning the evaluator found significant issues based on the ten evaluation criteria. A score of 0.5 indicates a neutral response, reflecting a balance of positive and negative observations. A score of 1 represents a positive evaluation. This evaluation was conducted by four experts for two documents—Document A and Document B—with each document being evaluated by two experts.

\section{Conclusion and Future Work}
Summarization plays a crucial role in our technological lives today, especially in the research field, where researchers must review numerous studies to gain insights and understand their content. To save time, automated summarization proves to be a necessary and important tool in this area. This work presents an ATS model for Kurdish Sorani research documents. The data includes research written in this language, collected from four departments, both directly and from their associated journals, totaling 231 documents. In our model's methodology, preprocessing steps were conducted, including segmentation and feature extraction. The model then employs two algorithms, Sentence Weighting and TF-IDF, to generate an extracted summary. We trained our data in two experiments: The first experiment included research documents with their conclusions. The second experiment excluded the conclusions from the research documents. The study includes both manual and automated ROUGE evaluations by comparing the research abstracts with our generated summaries. We also compared the results from the two experiments. Our study developed an ATS model capable of performing domain-specific summarization for research written in Sorani Kurdish. As a result of our experiments, we obtained average accuracy results with the conclusions removed from the research documents. Finally, our best result was a ROUGE-1 score of 19.58\%. To improve the performance and accuracy of Automatic Text Summarization (ATS) for Kurdish research documents, several enhancements can be considered. Incorporating additional algorithms into the model could enhance summary quality. Expanding the dataset and applying the model to diverse domains, along with ensuring clean, accurate data, would further improve performance. Experimenting with different data splits may also yield better results. Future work could involve training the model on various text types beyond research papers and extending it to other Kurdish dialects. Improving preprocessing is crucial, as Kurdish lacks robust tools compared to high-resource languages. Finally, transitioning from extractive to abstractive summarization could produce more natural, human-like summaries. These steps can help advance ATS systems for Kurdish texts.

\section*{Acknowledgment}
We would like to express our sincere gratitude to Lecturer Mr. Hawkar Jabbar HamadAli for his invaluable support with data collection, reaching out to manual evaluators, and participating as a manual evaluator in the role of a writer. We also thank Mr. Saman Idress Mahmood, MSc graduate from the University of Kurdistan Hawler, for assisting with converting our data from PDF to text. We are deeply grateful to the experts who contributed to the manual evaluation process: Asst. Prof. Barzan Jawhar Sadiq (Salahaddin University, Political Science Department), Dr. Azad Aziz Sleman (Salahaddin University, Kurdish Department), Dr. Awara Kamal Salih (Erbil Polytechnic University, Kurdish Department), Dr. Awat Ahmad Mohammed Salih (Lebanese French University, Kurdish Department), and Dr. Hazhar Ahmad Abdulghafur (Soran University, Kurdish Department). We also appreciate the support of the staff at the General Library of Salahaddin University, especially Ms. Snoor Balani, for providing the necessary access credentials. Additionally, we are thankful to Dr. Ali Abbas Mahmood for his assistance with proofreading this thesis.

\bibliographystyle{lrec}

\bibliography{KurdishATS}

\end{document}